# Noise Influence on the Fuzzy-Linguistic Partitioning of Iris Code Space

I.M. Motoc[1], C.M. Noaica[1], R. Badea[1], C.G. Ghica[2]

*Abstract*— This paper analyses the set of iris codes stored or used in an iris recognition system as an f-granular space. The f-granulation is given by identifying in the iris code space the extensions of the fuzzy concepts *wolves*, *goats*, *lambs* and *sheep* (previously introduced by Doddington as '*animals*' of the biometric menagerie) – which together form a partitioning of the iris code space. The main question here is how objective (stable / stationary) this partitioning is when the iris segments are subject to noisy acquisition. In order to prove that the f-granulation of iris code space with respect to the fuzzy concepts that define the biometric menagerie is unstable in noisy conditions (is sensitive to noise), three types of noise (*localvar, motion blur, salt and pepper*) have been alternatively added to the iris segments extracted from University of Bath Iris Image Database. The results of 180 exhaustive (all-to-all) iris recognition tests are presented and commented here.

I. INTRODUCTION

THIS paper assumes that the set of iris codes stored or used in an iris recognition system is an f-granular [12] space. The f-granules of this space are collections of iris codes identified as extensions of the fuzzy concepts *wolves, goats, lambs* and *sheep* introduced by Doddington (in speech recognition, [3]) and Yager (in iris recognition, [11]) as '*animals*' of the biometric menagerie, which together form a partitioning of the iris code space (further denoted as ICS). Let us denote this partitioning as Fuzzy Biometric Menagerie (FBM, [9]). Is FBM an objective (stable / stationary) attribute of the group of persons that pass through different iris recognition systems functioning in different calibration regimes? Different iris recognition system may have different security levels (is not the same thing to use a safety threshold and to use a safety band for separating genuine and imposter comparisons, for example), different procedures for encoding the iris texture, may use iris codes of different size and iris images acquired in noisy conditions. Is FBM invariable when any of these calibration parameters change? This question partially got its answer in [9] where Popescu-Bodorin et al.

[1] IEEE Student Member (Artificial Intelligence & Computational Logic Lab., Mathematics & Computer Science Department, Spiru Haret University, Bucharest, Romania), Email: motoc, noaica, badea [at] irisbiometrics.org
[2] IEEE Student Member (Artificial Intelligence & Computational Logic Lab., Mathematics & Computer Science Dept., Spiru Haret University, Bucharest, Romania), Programmer at Clintelica AB, claudiu.ghica [at] clintelica.com.

have shown that a change of the iris texture encoder, of the iris code size, or of the security settings determines a change of the FBM partitioning.

As a continuation of the research undertaken in [9], this paper demonstrates that noisy acquisition procedures also change the FBM partitioning of ICS.

In order to prove that a noisy acquisition change the FBM partitioning of ICS, 180 exhaustive iris recognition tests were performed using iris codes of dimensions 64x4, 128x8 and 256x16, obtained from the unwrapped iris segments that were randomly artificially noised with *localvar, salt and pepper* and *motion blur*. The tests were performed using two security settings (EER threshold and a safety band) that allow us to identify the *extensions of the f-concepts wolf / goat*. In order to highlight that the FBM partitioning of the ICS is unstable, the experimental results of our tests were compared with those obtained in [9], also.

II. FBM VS SYSTEM LOGIC AND SAFETY MODELS

All templates stored in an iris recognition system are *sheep-templates* if and only if, there is a clear gap between the maximum imposter similarity score and the minimum genuine similarity score. In other words, the *sheep-templates* are those validating a *consistent theory* of iris recognition like the ones formalized in [8] as:

$$\mathcal{P}_C \rightarrow [max\{S(C)|C \in \mathcal{I}\} \lneq min\{S(C)\}|C \in \mathcal{G}],$$

$$\mathcal{P}_C \rightarrow [max\{S(C)|C \in \mathcal{I}\} \ll min\{S(C)|C \in \mathcal{G}\}],$$

where $\mathcal{P}_C$ denotes "*a conjunction of prerequisite conditions (relative to the image acquisition and processing at all levels from eye image to the iris code) expressed in binary logic*" (as said in [8]), $C = \mathcal{I} \cup \mathcal{G}$ is the natural disjoint partitioning of comparison space in imposter and genuine pairs, $S$ is the similarity score and $C$ denotes a comparison.

The fact that experimental genuine and imposter score distributions do not overlap each other certifies that no impersonation occurred in the system (there is no support for a false accept) and no user matches himself so bad such that to generate a false reject. Therefore, in such a case the extensions (the referents) of the concepts '*wolves*', '*goats*' and '*lambs*' are empty, whereas the extension of the concept '*sheep*' is the entire set of iris codes recorded / tested in the biometric system. Fig. 7.c from [2] illustrates such an example. As seen there, in an example like that, the pessimistic envelopes of the imposter and genuine score distributions may help one identifying templates (or users) that are good candidates for the roles of lamb, wolf or goat, in a future in which the number of users or at least the number of templates stored in the database grows. This perspective is very well described by a Fuzzy 3-Valent Disambiguated Model (F3VDM, [6]) of biometric security. The following figure illustrates the correspondence between the FBM and the F3VDM associated to the iris recognition tests undertaken in [2] for the dual iris combined HH&LG Encoder (see also Fig. 7.c, Fig. 7.d, Fig. 8, Table 2 in [2]) approach:

| I | [0.6050, 1] | Genuine Pairs | False Accept Rate: POFA(0.6050) = 1.0777E-10<br>True Accept Safety: 1-POFA(0.6050) | |
|---|---|---|---|---|
| O | (0.5800, 0.6050) | Artificially Undecidable Pairs | Genuine Discomfort Rate <<br>POFR(0.6050) ≈ 6.1282E-4<br>Imposter Discomfort Rate <<br>POFA(0.5800) ≈ 9.7924E-7<br>---<br>Total Discomfort Rate ≈ **6.1379E-4** | INCERTITUDE / DISCOMFORT: CANDIDATE: WOLVES, LAMBS, GOATS / SECURITY: ACTUAL SHEEP |
| D | [0, 0.5800] | Imposter Pairs | False Reject Rate: POFR(0.5800) = 1.09E-4<br>False Reject Safety: 1-POFR(0.5800) | |

Fig. 1. An illustration of the relation between the Fuzzy Biometric Menagerie and the Fuzzy 3-Valent Disambiguated Model of biometric security.

On the contrary, an overlapping between the genuine and imposter distributions certifies that there are indeed ambiguous scores obtained in the system, scores for which an imposter pair (comparison) could be labeled as being a genuine pair (comparison) or vice versa, depending on what threshold the system would use for giving the biometric decision. For example, if the score distributions overlap each other and the threshold is under the minimum genuine score, we could talk about goats in terms of candidates and about wolves and lambs as exemplified certitudes. If the score distributions overlap each other and the threshold is above the maximum imposter score, we could talk about wolves and lambs in terms of candidates and about goats as exemplified certitudes. When the distributions overlap each other and the security threshold is strictly between the minimum genuine and the maximum imposter score, all the concepts wolves, lambs and goats are exemplified.

### III. EXPERIMENTAL RESULTS

All the tests undertaken in this paper relay on the University of Bath Iris Image Database (UBIID, [10]). Circular Fuzzy Iris Segmentation procedure (proposed in [4], [5], available for download in [6]) facilitates the extraction of the unwrapped iris segments for the iris image available in the database. In order to extract the iris segments characteristics we used two encoders: Log-Gabor and a modified version of Haar-Hilbert [5] (the version from [6] has a limitation with respect to iris code size).

As seen in [9], the partitioning of the iris code space as a Biometric Menagerie is fuzzy and not quite objective. In other words, Fuzzy Biometric Menagerie is sensitive to system calibration variables. This article analyzes if the Fuzzy Biometric Menagerie is sensitive also to noise, or not. We consider here the cases in which artificial noise (localvar, motion blur and salt and pepper) is added to the initial iris segments.

The purpose of the tests was to realize a FBM partitioning of ICS while using a threshold and a safety band for each of the iris segment dimensions and for each type of artificial noise added (using the same noise intensity for the iris segments of the same dimension). Noise parameters were set such that the EER values of our tests to be double (at most) than those obtained in [9] for iris recognition tests that did not use noise. The *marginal* and the *last wolf-* and *goat-templates* (introduced in [9]) identified in our tests were also compared with the ones obtained in [9] to show that the FBM partitioning is unstable and its instability is influenced by noise (artificial noise in this case). The total number of exhaustive iris recognition tests undertaken here is 180. A series of five tests has been done for each type of noise, iris code dimension, encoder used and security setting.

Because of the space restrictions that we must respect here, this paper presents only a selection of experimental results corresponding to 180 exhaustive all-to-all iris recognition tests described in the technical report [1].

The first half of this section presents results obtained after using a safety band that was narrowed until *marginal/first wolf-* and *goat-templates* [9] were found, while the second half presents *last wolf-* and *goat-templates*, i.e. results obtained after using the EER threshold ($t_{EER}$).

#### A. Experimental results obtained for safety bands

As in [9], the safety bands used here are determined by narrowing the maximal safety band [mGS, MIS] until goat- and wolf-templates appear in the system, meaning that the concepts of *wolf-* and *goat-templates* refer non-empty sets of iris codes. More exactly, the templates determined in this section are the *first wolf-* and *goat-templates* obtained for the first safety band that allow them to exist.

Fig. 2 presents the behavior of four *marginal wolf-templates* obtained for Haar-Hilbert and Log-Gabor encoders, and illustrates the fact that the number of impersonations could increase along with the increase of the iris code dimension.

Fig. 2.a and Fig. 2.b present the similarity scores obtained for the *first wolf-templates* (determined as iris codes of dimension 64x4). By comparing them, we can notice that the number of impersonations associated to *first wolf-templates* could differ from one encoder to another.

For iris codes of dimension 128x8, the *first wolf-template* with the highest number of impersonation was the one obtained for Haar-Hilbert encoder, as seen in Fig. 2.c and Fig. 2.d.

Table 1 presents the *marginal wolf-templates* obtained for Haar-Hilbert encoder when using iris codes of dimension 64x4 after performing 15 tests (3 series of 5 tests, each series with a different type of noise). The noise introduced in the unwrapped iris segments clearly influences the FBM partitioning of ICS. Consequently, almost all marginal wolf-templates detected in our tests differ from that obtained and presented in [9]. More than that, the templates obtained here for the same noise and same intensity are different, which means that the marginal wolf-templates depend also on the randomness of the noise.

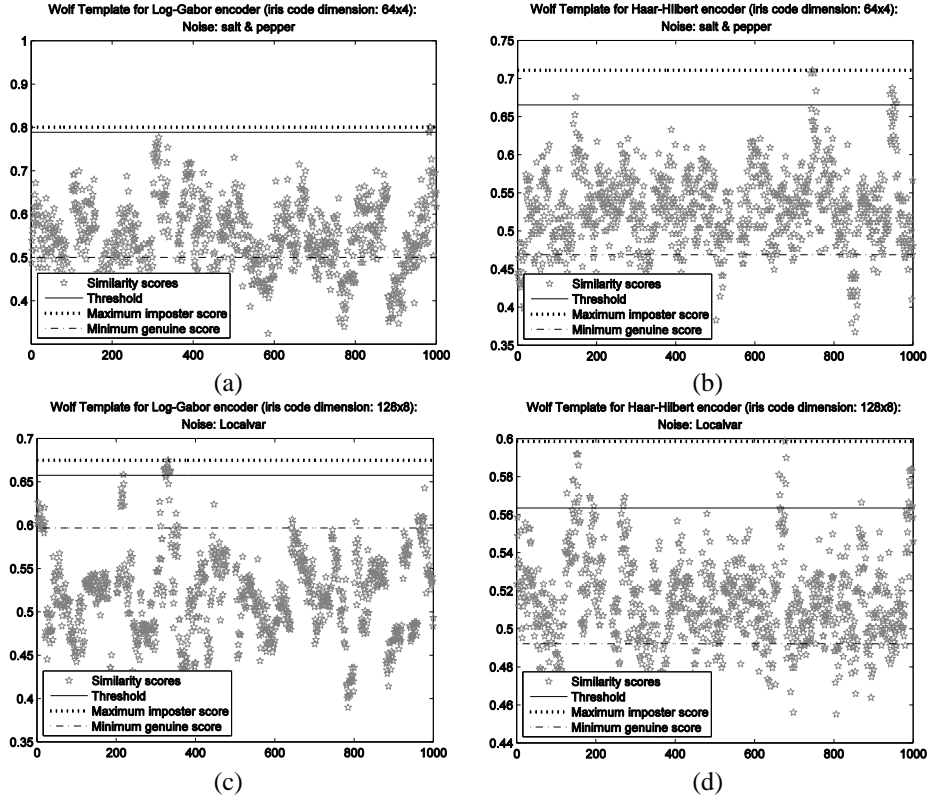

Fig. 2. The similarity scores associated to the imposter comparisons generated by the *marginal wolf-templates* along four tests: Log-Gabor (64x4, salt and pepper, 357 - a; 128x8, localvar, 482 - c) and Haar-Hilbert (64x4, salt and pepper, 546 - b, 128x8, localvar, 227- d) encoders

TABLE I
*MARGINAL WOLF-TEMPLATES* OBTAINED FOR HAAR-HILBERT ENCODER (64X4)

|  | Impersonations | Template | Safety band | Width |
|---|---|---|---|---|
| Images without noise [9] | 15 | 549 | [0.6091, 0.6722] | 0.0631 |
| Salt and pepper | 4, 8, 3, 3, 10 | 503, 165, 88, 230, 546 | [0.5044, 0.6987], [0.5656, 0.6648], [0.5127, 0.6944], [0.5317, 0.7066], [0.5338, 0.6655] | 0.1943, 0.0992, 0.1817, 0.1749, 0.1317 |
| Motion Blur | 3, 4, 3, 4, 3 | 239, 387, 959, 807, 505 | [0.5630, 0.6987], [0.5595, 0.6905], [0.5741, 0.6994], [0.5523, 0.7016], [0.5534, 0.7044] | 0.1357, 0.1310, 0.1253, 0.1493, 0.1510 |
| Localvar | 3, 3, 3, 3, 4 | 541, 501, 510, 926, 755 | [0.5791, 0.7139], [0.5864, 0.7183], [0.5708, 0.7144], [0.5117, 0.7061], [0.5584, 0.7072] | 0.1348, 0.1319, 0.1436, 0.1944, 0.1488 |

TABLE II
*MARGINAL WOLF-TEMPLATES* OBTAINED FOR LOG-GABOR ENCODER (128X8)

|  | Impersonations | Template | Safety band | Width |
|---|---|---|---|---|
| Images without noise [9] | 17 | 484 | [0.6277, 0.6555] | 0.0278 |
| Salt and pepper | 9, 7, 14, 12, 8 | **481**, **481**, 482, **481**, **481** | [0.6206, 0.6695], [0.6197, 0.6645], [0.6197, 0.6527], [0.6227, 0.6596], [0.6216, 0.6714] | 0.0489, 0.0448, 0.0330, 0.0369, 0.0498 |
| Motion Blur | 11, 10, 9, 19, 17 | **481**, **481**, **481**, 485, 484 | [0.6158, 0.6654], [0.6258, 0.6671], [0.6039, 0.6735], [0.6163, 0.6415], [0.6114, 0.6484] | 0.0496, 0.0413, 0.0696, 0.0252, 0.0370 |
| Localvar | 14, 14, 14, 14, 14 | **482**, **482**, **482**, **482**, **482** | [0.6267, 0.6575], [0.6267, 0.6575], [0.6267, 0.6575], [0.6267, 0.6575], [0.6267, 0.6575] | 0.0308, 0.0308, 0.0308, 0.0308, 0.0308 |

Table 2 presents the *marginal wolf-templates* as iris codes of dimension 128x8 obtained using Log-Gabor encoder. The templates identified here as *marginal wolf-templates* are different from that obtained in [9] for the same encoder (see Table 3 from [9]). This fact supports the idea that the *marginal wolf-template* depends on the noise. Table 2 illustrates that, for *localvar* noise, the *marginal wolf-templates* are identical to each other, which leads to the idea that the randomness of this type of noise do not affect any of the results (template, number of impersonations, safety band and its width). For *salt and pepper* and *motion blur*, the templates differ in three out of ten cases, which mean that the result could depend on the type of noise.

The comparison between Fig. 3.a - Fig. 3.b and Fig. 3.c – Fig. 3.d illustrates that, along with the increasing dimension of the iris code, the number of mismatches corresponding to the *marginal goat-templates* can decrease considerably when the iris codes are generated using Log-Gabor encoder.

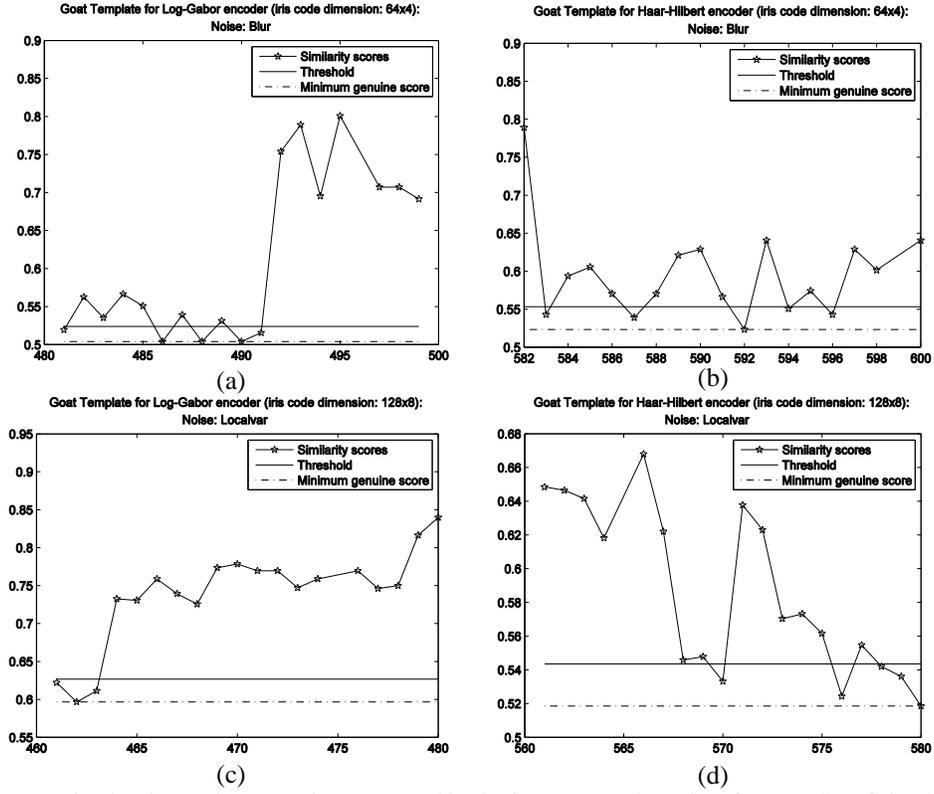

Fig. 3. The similarity scores associated to the genuine comparisons generated by the *first goat-templates* along four tests: Log-Gabor (64x4, blur, 496 - a; 128x8, localvar, 475 - c) and Haar-Hilbert (64x4, blur, 581 - b, 128x8, localvar, 565 - d) encoders

TABLE III
*MARGINAL GOAT-TEMPLATES* OBTAINED FOR HAAR-HILBERT ENCODER (64X4)

|  | Impersonations | Template | Safety band | Width |
|---|---|---|---|---|
| Images without noise [9] | 3 | **565** | [0.6091, 0.6722] | 0.0631 |
| Salt and pepper | 2, 2, | **580**, 566, | [0.5044, 0.6987], [0.5656, 0.6648], | 0.1943, 0.0992, |
|  | 2, 2, | 581, **565**, | [0.5127, 0.6944], [0.5317, 0.7066], | 0.1817, 0.1749, |
|  | 2 | **580** | [0.5338, 0.6655] | 0.1317 |
| Motion Blur | 2, 2, | 566, **581**, | [0.5630, 0.6987], [0.5595, 0.6905], | 0.1357, 0.1310, |
|  | 2, 3, | 597, **565**, | [0.5741, 0.6994], [0.5523, 0.7016], | 0.1253, 0.1493, |
|  | 5 | **581** | [0.5534, 0.7044] | 0.1510 |
| Localvar | 2, 2, | 582, 566, | [0.5791, 0.7139], [0.5864, 0.7183], | 0.1348, 0.1319, |
|  | 2, 2, | **581**, **565**, | [0.5708, 0.7144], [0.5517, 0.7061], | 0.1436, 0.1544, |
|  | 2 | **581** | [0.5584, 0.7072] | 0.1488 |

TABLE IV
*MARGINAL GOAT-TEMPLATES* OBTAINED FOR HAAR-HILBERT ENCODER (128X8)

|  | Impersonations | Template | Safety band | Width |
|---|---|---|---|---|
| Images without noise [9] | 3 | **565** | [0.5456, 0.6823] | 0.1367 |
| Salt and pepper | 2, 2, | 580, **565**, | [0.5198, 0.5925], [0.5256, 0.5955], | 0.0727, 0.0699, |
|  | 2, 2, | 580, 580, | [0.5336, 0.5963], [0.5278, 0.5864], | 0.0627, 0.0586, |
|  | 2 | 580 | [0.5307, 0.5884] | 0.0577 |
| Motion Blur | 2, 2, | 580, 580, | [0.5287, 0.5983], [0.5365, 0.5944], | 0.0767, 0.0579. |
|  | 2, 2, | 580, 580, | [0.5237, 0.6004], [0.5256, 0.5945], | 0.0767, 0.0689, |
|  | 2 | 580 | [0.5178, 0.6013] | 0.0835 |
| Localvar | 5, 2, | **565**, **565**, | [0.5436, 0.5863], [0.5228, 0.5885], | 0.0427, 0.0657, |
|  | 2, 2, | **565**, 580, | [0.5322, 0.5635], [0.5162, 0.5766], | 0.0313, 0.0604, |
|  | 2 | **565** | [0.5188, 0.5935] | 0.0747 |

The discomfort rate could decrease along with the increase of the iris code dimension, as seen in Fig. 3.c and Fig. 3.d.

As seen in Table 3, for each noise, in one out of five tests, the *marginal goat-template* was the same with that presented in Table3 from [9]. This can happen from two reasons: (1) the Haar wavelet transform was able to remove a significant part of the noise, and (2) the noise influence can be insignificant enough to have the same goat-template. For each noise, the templates were different in 3 out of 5 successive tests, performed at the same noise intensity, which means that the randomness of the noise could change the *marginal goat-template*.

Table 4 illustrates that 5 out of 15 *marginal goat-templates* obtained in our tests are identical with that presented in [9], the reasons being the same as the ones presented above. Even if in the majority of our tests the *marginal goat-templates* were different from the one presented in [9], the templates were identical for all three noises.

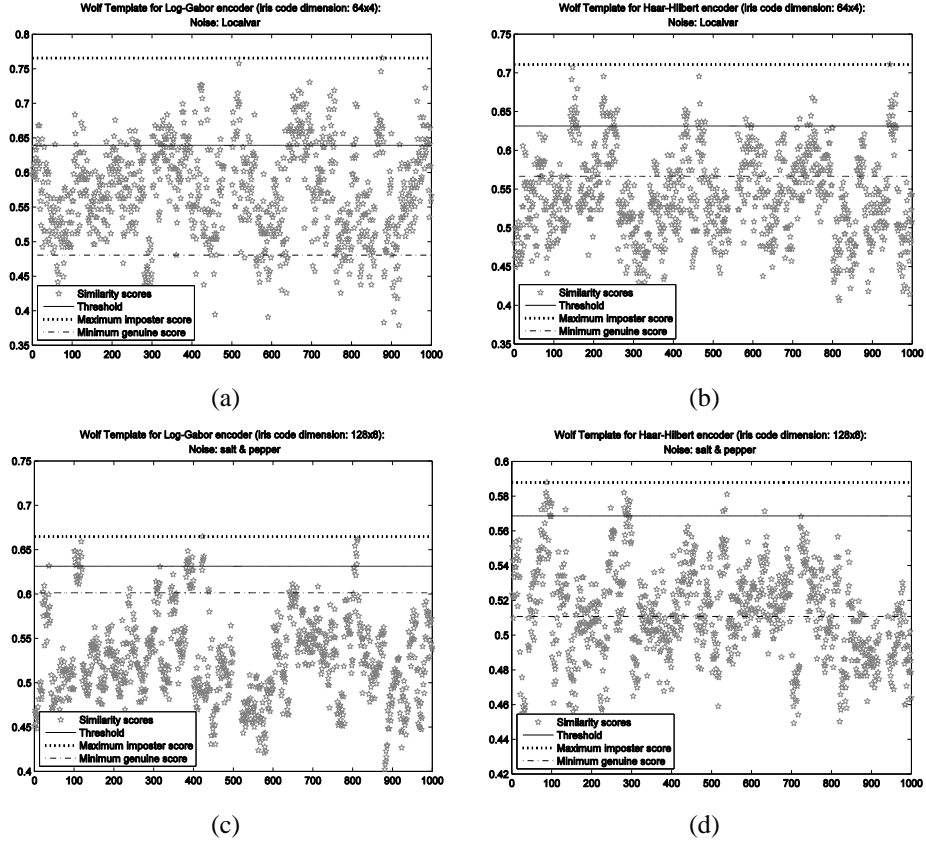

Fig. 4. The similarity scores associated to the imposter comparisons generated by the *last wolf-templates* along four tests: Log-Gabor (64x4, localvar, 495 - a; 128x8, salt and pepper, 505 - c) and Haar-Hilbert (64x4, localvar, 548 - b, 128x8, salt and pepper, 924 - d) encoders

*B. Experimental results obtained for $t_{EER}$ thresholds*

This subsection presents the results obtained in 60 selected exhaustive iris recognition tests performed by running the biometric system at EER threshold $t_{EER}$. The intensity of the noise in all of these tests was limited by the condition that for a given encoder and for a given code dimension, the EER values obtained in our test to be at most double than the values obtained in [9].

TABLE V
*LAST WOLF-TEMPLATES* OBTAINED FOR LOG-GABOR ENCODER (64X4)

|  | Imperson ation | Template | EER | $t_{EER}$ |
|---|---|---|---|---|
| Images without noise [9] | 63 | 236 | 4.08E-2 | 0.7529 |
| Salt and pepper | 134, 156, 147, 137, 135 | 679, 250, 382, 240, 250 | 4.27E-2, 4.55E-2, 4.82E-2, 4.37E-2, 4.28E-2 | 0.6471, 0.6431, 0.6431, 0.6431, 0.6471 |
| Motion blur | 147, 147, 147, 171, 139 | 679, 669, 389, 389, 240 | 4.29E-2, 4.32E-2, 4.16E-2, 4.37E-2, 4.11E-2 | 0.6431, 0.6431, 0.6431, 0.6431, 0.6431 |
| Localvar | 134, 183, 163, 148, 169 | 238, 495, 387, 117, 492 | 4.18E-2, 4.51E-2, 4.56E-2, 4.54E-2, 4.57E-2 | 0.6431, 0.6392, 0.6392, 0.6392, 0.6392 |

Fig. 4.a and Fig. 4.b represent the behavior of the *last wolf-templates* under the influence of *localvar*, showing that the number of impersonations is higher for the template obtained for Log-Gabor than the one obtained for Haar-Hilbert. On the contrary, for the *last wolf-templates* represented in Fig. 4.c and Fig. 4.d, the number of impersonations is higher for the one obtained for Haar-Hilbert encoder.

Table 5 presents the experimental results obtained in [9] and in our iris recognition tests when searching for *last wolf-templates*. All tests use iris codes of dimension 64x4 generated with Log-Gabor iris texture encoder. For each artificial noise, the *last wolf-templates* detected in our tests are different from the one mentioned in Table 3 from [9] and, most of them, are even different from each other. By comparing our results with those obtained in [9], for the same experimental setup (except the presence of the noise), it can be observed that the number of impersonations obtained in [9] is significantly smaller than the ones obtained in our tests (the presence of noise stimulates impersonation). The EER points from our tests are different from each other (with one exception, when the EER value is 4.37E-2 for the fourth test in both *salt and pepper* and *motion blur*) and different from those presented in [9]. These results advocate for the subjectivity of "*last wolf-template*" concept.

Table 6 stores the experimental results obtained in those tests in which the Log-Gabor iris texture encoder generates iris codes of dimension 128x8. By comparison, the *last wolf-templates* identified in our tests are all different from the one presented in [9], and in general different from each other, also This highlights the fact that the noise is a very important factor that affects the FBM partitioning of ICS. By analyzing the *last wolf-templates* resulted from our tests for each noise, we found noticeable that the extension of the *f-concept wolf* changes from one test to another.The tests using the *localvar*

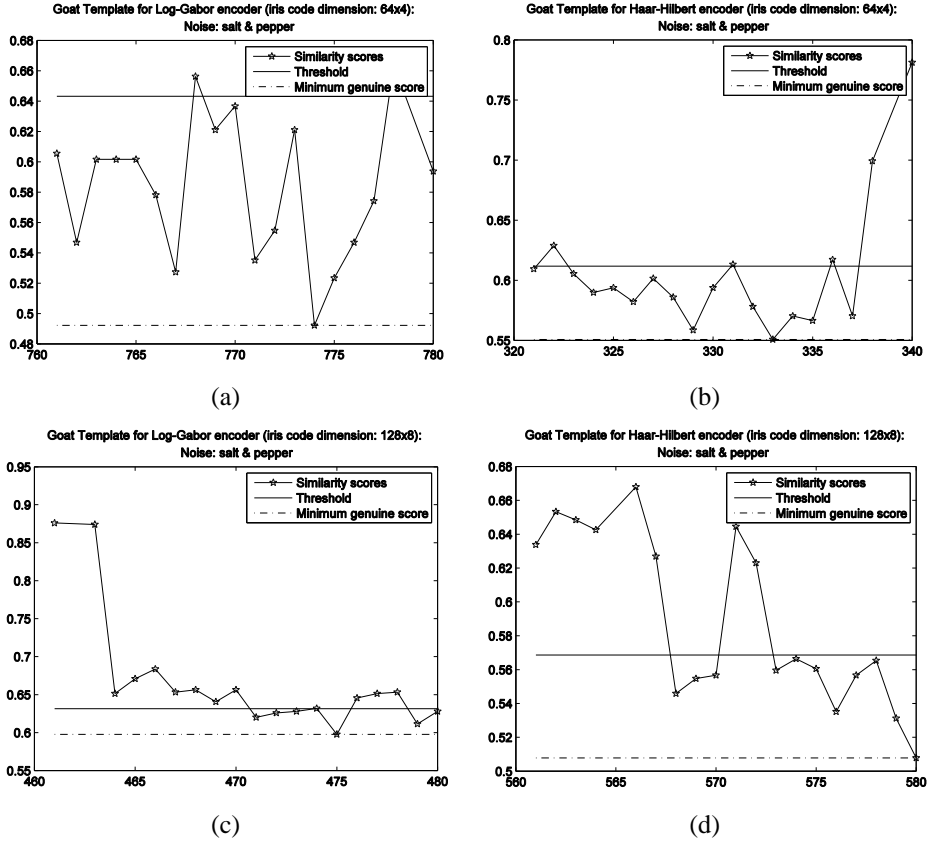

Fig. 5. The similarity scores associated to the genuine comparisons generated by the *last goat-templates* along four tests: Log-Gabor (64x4, salt and pepper, 779 - a; 128x8, salt and pepper, 462 - c) and Haar-Hilbert (64x4, salt and pepper, 339 - b, 128x8, salt and pepper, 565 - d) encoders

noise are the only ones that provide identical *last wolf-templates*. However, this is an isolated case that could not advocate for the objectivity of the '*wolf*' concept.

TABLE VI
*LAST WOLF TEMPLATES* OBTAINED FOR LOG-GABOR ENCODER (128X8)

| | Impersonations | Template | EER | $t_{EER}$ |
|---|---|---|---|---|
| Images without noise [9] | 22 | 392 | 9.37E-4 | 0.6392 |
| Salt and pepper | 25, 24, 29, 22, 29 | **387**, 395, **387**, 484, **505** | 1.70E-3, 1.30E-3, 1.80E-3, 1.80E-3, 1.70E-3 | 0.6314, 0.6353, 0.6275, 0.6314, 0.6314 |
| Motion blur | 23, 24, 27, 21, 28 | **387**, 485, **505**, 481, 118 | 1.40E-3, 1.70E-3, 1.30E-3, 1.40E-3, 1.80E-3 | 0.6353, 0.6353, 0.6353, 0.6353, 0.6314 |
| Localvar | 20, 20, 20, 20, 20 | **482**, **482**, **482**, **482**, **482** | 1.10E-3, 1.10E-3, 1.10E-3, 1.10E-3, 1.10E-3 | 0.6392, 0.6392, 0.6392, 0.6392, 0.6392 |

Fig. 5.a and Fig. 5.b represent the behavior of the *last goat-templates* resulted for iris codes of dimension 64x4 generated under the influence of *salt and pepper* noise. Between the two presented templates, the one encoded with Haar-Hilbert has the highest number of rejections (17) which means that the noise distribution affected this template more than the one obtained for Log-Gabor. The similarity scores oscillate more in Fig. 5.a than in Fig. 5.b. The same thing happens in Fig. 5.c and Fig. 5.d. This shows that Haar-Hilbert encoder is much more sensitive to noise influence.

Table 7 presents the case when one of the *last goat-templates* obtained in our tests is identical with the one presented in [9]. However, this is an isolated case, also.

TABLE VII
LAST GOAT-TEMPLATES OBTAINED FOR LOG-GABOR ENCODER (64X4)

| | Rejections | Template | EER | $t_{EER}$ |
|---|---|---|---|---|
| Images without noise [9] | 11 | **493** | 4.08E-2 | 0.7529 |
| Salt and pepper | 12, 18, 15, 17, 18 | **493**, 993, 359, **779**, 443 | 4.27E-2, 4.55E-2, 4.82E-2, 4.37E-2, 4.28E-2 | 0.6471, 0.6431, 0.6431, 0.6431, 0.6471 |
| Motion blur | 14, 15, 17, 12, 15 | **360**, 359, **779**, 817, 360 | 4.29E-2, 4.32E-2, 4.16E-2, 4.37E-2, 4.11E-2 | 0.6431, 0.6431, 0.6431, 0.6431, 0.6431 |
| Localvar | 13, 15, 14, 16, 16 | 359, 429, 359, **359**, 119 | 4.18E-2, 4.51E-2, 4.56E-2, 4.54E-2, 4.57E-2 | 0.6431, 0.6392, 0.6392, 0.6392, 0.6392 |

TABLE VIII
LAST GOAT-TEMPLATES OBTAINED FOR HAAR-HILBERT ENCODER (128X8)

| | Rejections | Template | EER | $t_{EER}$ |
|---|---|---|---|---|
| Images without noise [9] | 6 | 565 | 1.70E-3 | 0.5765 |
| Salt and pepper | 9, 8, 8, 11, 11 | 565, 565, 565, 565, 565 | 2.70E-3, 2.60E-3, 2.70E-3, 2.80E-3, 2.60E-3 | 0.5686, 0.5686, 0.5686, 0.5686, 0.5686 |
| Motion blur | 8, 8, 9, 8, 10 | 565, 565, 565, 565, 565 | 2.20E-3, 2.00E-3, 2.10E-3, 2.10E-3, 2.30E-3 | 0.5765, 0.5765, 0.5765, 0.5765, 0.5765 |
| Localvar | 9, 10, 9, 8, 8 | 565, 565, 565, 565, 565 | 3.40E-3, 3.10E-3, 3.10E-3, 2.60E-3, 3.40E-3 | 0.5647, 0.5647, 0.5647, 0.5686, 0.5647 |

Table 8 stores the experimental results obtained using iris codes of dimension 128x8. All our tests in this series indicate the same *last goat-template*, which is the same template obtained in [9], as well. After a visual examination of the eye image that corresponds to the *last goat-template* detected (565), and after a visual comparison of the pairs of images associated to the genuine goat scores, we found that: firstly, the subject wear contact lens (the contact lenses can damage the segmentation process performance), and secondly, the limbic boundary detected for the image 565 was misplaced accidentally, and therefore the physical support indicated by it is not the actual limbic boundary. This situation highlights the fact that the FBM partitioning of ICS is depending on the segmentation process, also.

All the 120 experimental results presented in this section show that the noise is an important factor that influences the FBM partitioning of ICS. There have been a few cases when the template was candidate to be a *last wolf-/goat-template* several consecutive times but these cases are isolated, or there is a strong and objective reason for them to happen (as the one presented in Table 8). In conclusion, the instability of the identified wolf / goats templates indicates that a noisy acquisition process is an important factor that could influence the performances of an iris recognition system.

## IV. CONCLUSIONS

This paper shown that the FBM partitioning of iris code space, the consistency and fuzziness of FBM and of its underlying concepts all depend not only on the system calibration (in terms of iris texture encoder and iris code dimension), but also on the noise that could affect iris image acquisition process. The experimental results from a total of 36 series of iris recognition tests (5 tests in each series) undertaken for Bath Database shown that, in iris recognition, the so-called Biometric Menagerie definitely is a fuzzy and inconsistent concept. The extensions of the fuzzy concepts '*wolf*' and '*goat*' vary under the influence of noisy acquisition and system calibration.